\documentclass[10pt,twocolumn,letterpaper]{article}

\usepackage{cvpr}      
\usepackage{graphicx}
\usepackage{multirow}
\usepackage{algpseudocode} 
\usepackage{amsmath} 
\usepackage{amssymb}
\usepackage{wrapfig}
\usepackage{graphicx}
\usepackage{caption}
\usepackage{subcaption}
\usepackage{floatflt}
\usepackage{booktabs} 
\usepackage{array} 
\usepackage{xcolor}
\usepackage{lipsum}
\usepackage{colortbl}
\usepackage{algorithm}
\usepackage{amssymb}
\usepackage{float} 
\usepackage{afterpage}
\usepackage{stfloats}
\usepackage{multicol}
\usepackage{pifont}
\usepackage[T1]{fontenc}

\definecolor{cvprblue}{rgb}{0.21,0.49,0.74}
\usepackage[pagebackref,breaklinks,colorlinks,allcolors=cvprblue]{hyperref}

\def\paperID{3327} 
\def\confName{CVPR}
\def\confYear{2025}

\title{MoEE: Mixture of Emotion Experts for Audio-Driven Portrait Animation}

\author{
Huaize Liu\textsuperscript{1,2,5}, 
Wenzhang Sun\textsuperscript{2}, 
Donglin Di\textsuperscript{2},
Shibo Sun\textsuperscript{3},
Jiahui Yang\textsuperscript{3},
Changqing Zou\textsuperscript{4,1},\\
Hujun Bao\textsuperscript{4 \dag}
\\
\\
\textsuperscript{1}Zhejiang Lab,
\textsuperscript{2}Li Auto, 
\textsuperscript{3}Harbin Institute of Technology, 
\textsuperscript{4}Zhejiang University,\\
\textsuperscript{5}Hangzhou Institute for Advanced Study, University of Chinese Academy of Sciences
\\
}

\begin{document}


\maketitle

\renewcommand{\thefootnote}{\fnsymbol{footnote}}
\footnotetext[2]{Corresponding author.}

\begin{abstract}

The generation of talking avatars has achieved significant advancements in precise audio synchronization. 
However, crafting lifelike talking head videos requires capturing a broad spectrum of emotions and subtle facial expressions.
Current methods face fundamental challenges: 
\textbf{a)} the absence of frameworks for modeling single basic emotional expressions, which restricts the generation of complex emotions such as compound emotions; 
\textbf{b)} the lack of comprehensive datasets rich in human emotional expressions, which limits the potential of models.
To address these challenges, we propose the following innovations: 
\textbf{1)} the \textbf{M}ixture of \textbf{E}motion \textbf{E}xperts (\textbf{MoEE}) model, which decouples six fundamental emotions to enable the precise synthesis of both singular and compound emotional states; 
\textbf{2)} the DH-FaceEmoVid-150 dataset, specifically curated to include six prevalent human emotional expressions as well as four types of compound emotions, thereby expanding the training potential of emotion-driven models. 
Furthermore, to enhance the flexibility of emotion control, we propose an emotion-to-latents module that leverages multimodal inputs, aligning diverse control signals—such as audio, text, and labels—to ensure more varied control inputs as well as the ability to control emotions using audio alone.
Through extensive quantitative and qualitative evaluations, we demonstrate that the MoEE framework, in conjunction with the DH-FaceEmoVid-150 dataset, excels in generating complex emotional expressions and nuanced facial details, setting a new benchmark in the field. These datasets will be publicly released.
\end{abstract}

\section{Introduction}

\begin{figure}[h] 
    \centering
    \includegraphics[width=1.0\linewidth]{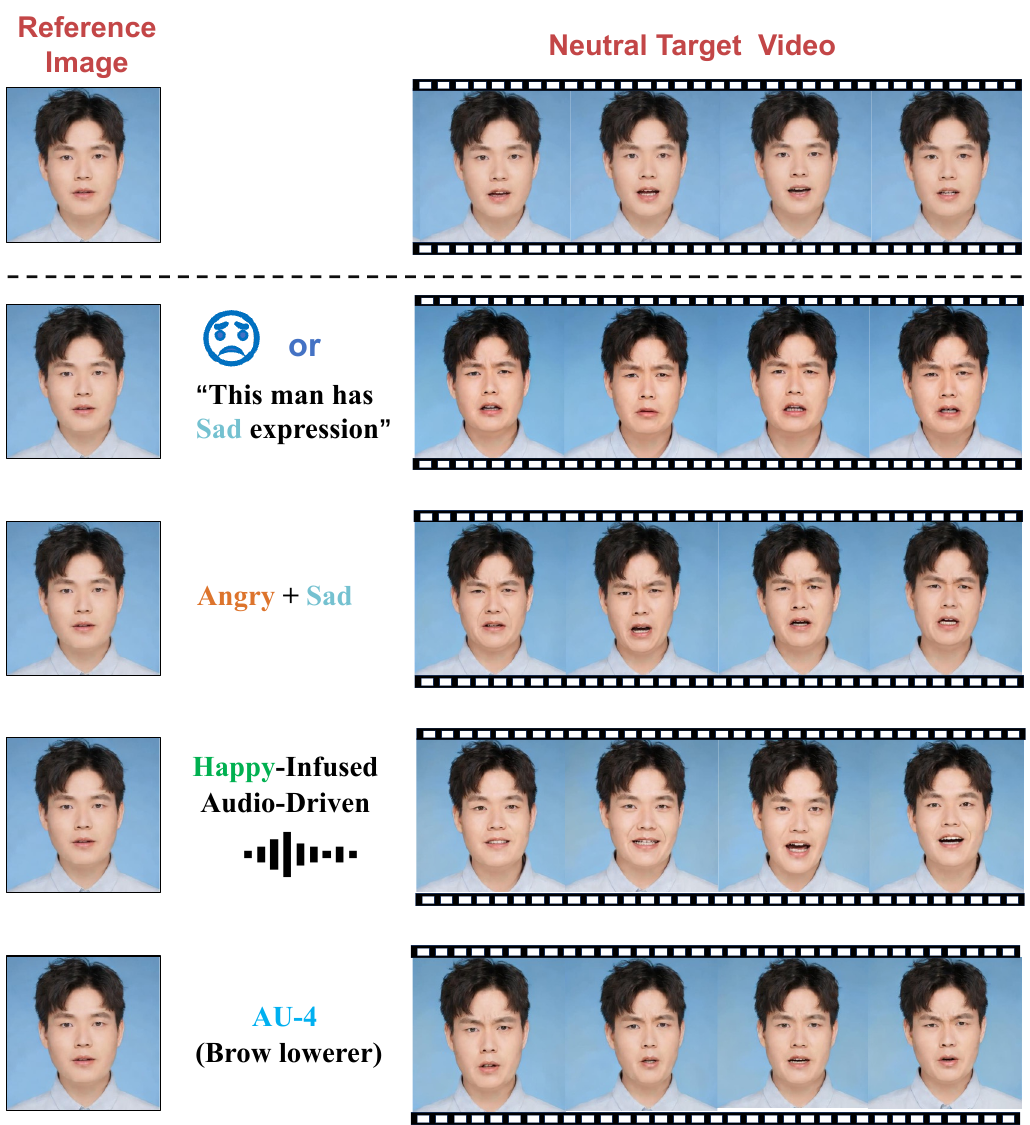} 
    \caption{MoEE enables more natural and vivid basic emotion control and compound emotion control, through labels or text prompts in the generated talking face. It can also achieve emotion control based solely on audio with emotional cues. Beyond coarse-grained emotion control (\eg, audio, label and text prompt), our method allows for fine-grained expression control through AU labels.}
    \label{fig:1_case}
    \vspace{-5mm}
\end{figure}

\begin{figure*}[!ht]
    \centering
    \includegraphics[width=\linewidth]{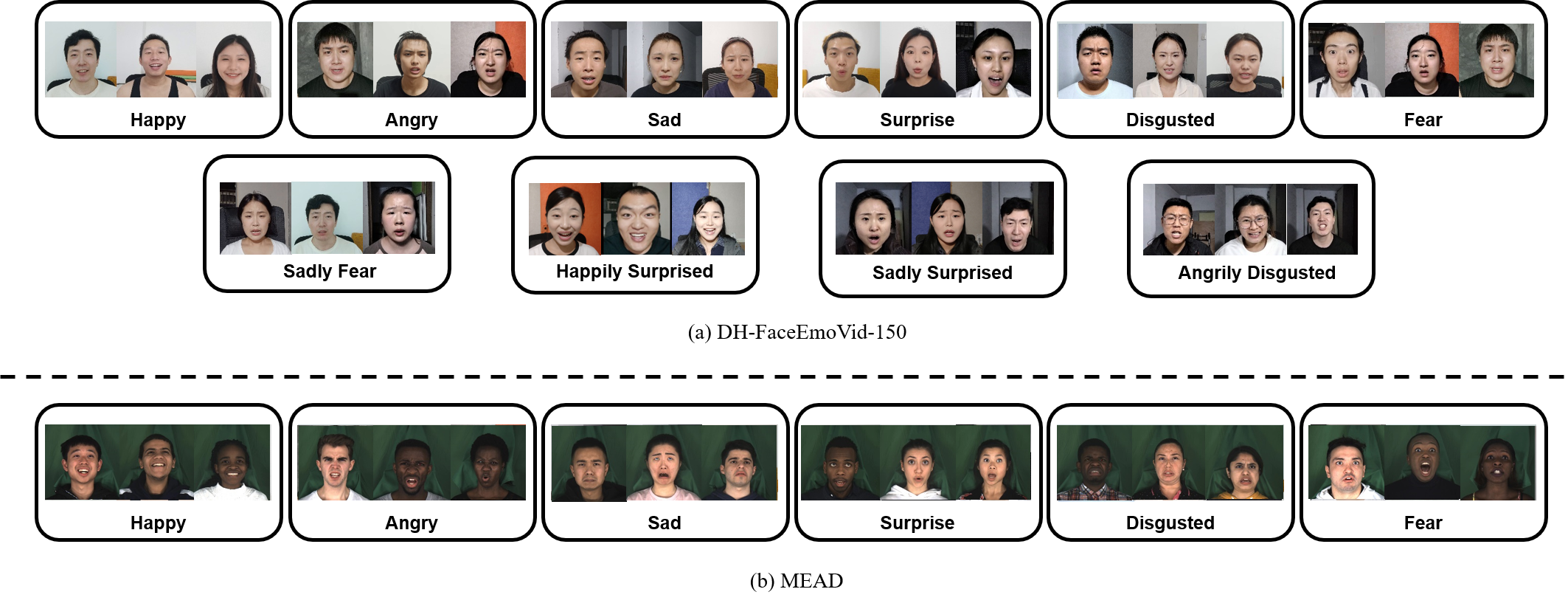}  
    \caption{Showcase of publicly available datasets and our proposed datasets: We refer to datasets like (a) DH-FaceEmoVid-150 and (b) MEAD as emotion datasets.}
    \label{fig:dataset1}
    \vspace{-5mm}
\end{figure*}

In recent years, audio-driven avatar generation has achieved significant advancements in precise audio synchronization \cite{chen2024echomimic, jiang2024loopy, wei2024aniportrait, xu2024hallo, zhang2023sadtalker, tian2024emo, liu2024anitalker, xu2024vasa}. However, creating realistic conversational videos requires more than just audio synchronization; it also demands the capture of a wide range of emotions and subtle facial expressions. Research in this area holds valuable applications not only in the entertainment industry but also in fields such as remote communication, virtual assistants, and mental health, showcasing considerable potential.

Previous studies attempt to integrate emotional and expressive information through labels, text, or videos \cite{sun2023vividtalk, ma2023styletalk, gan2023efficient, wang2024instructavatar, ma2023dreamtalk, emmn, zhai2023talking}. While these approaches can control basic emotions, they lack naturalness in generating single emotions and diversity in generating complex compound emotions. The fundamental limitations are twofold: first, the absence of precise frameworks for modeling basic emotions hinders accurate synthesis of complex emotional states. Second, the lack of comprehensive datasets rich in diverse human emotional expressions significantly limits the potential of these models to produce varied emotional outputs.

To address these issues, we draw inspiration from mixture of experts and propose the Mixture of Emotion Experts (MoEE) model. This model decouples six basic emotions, enabling precise synthesis of both singular and compound emotional states.
To push the performance boundaries of MoEE, we curate the DH-FaceEmoVid-150 dataset, comprising 150 hours of video content featuring six basic and four compound emotions. Initially, we fine-tune a referenceNet \cite{hu2023animateanyone,xu2024hallo} and a denoising unet \cite{rombach2022high,ho2020denoising} on the entire dataset. We then train six emotion expert networks with single emotion data for precise modeling, and a gating network with compound emotion data to synthesize complex expressions. This approach significantly boosts the generative quality and generalization ability of the MoEE model.

Furthermore, we design an emotion-to-latents module that aligns multimodal inputs (\ie, audio, text, labels, as illustrated in Figure~\ref{fig:1_case}) into a latent space using adaptive attention. Extensive quantitative and qualitative evaluations demonstrate that the MoEE framework, in conjunction with the DH-FaceEmoVid-150 dataset, excels in generating both singular and complex emotional expressions with nuanced facial details. Our contributions are as follows:

\begin{itemize}
    \item We present the Mixture of Emotion Experts (MoEE) model focused on facial emotional expression in audio-driven portrait animation. MoEE achieves lifelike generation quality for both singular and compound emotions with strong generalization ability.
    \item We introduce the DH-FaceEmoVid-150 dataset, a high-resolution (1080p) collection specifically for human facial emotional expressions. It comprises six basic and four compound emotions, providing multimodal information such as AU labels, text descriptions, and emotion categories for each video.
    \item We develop a module that aligns multimodal inputs to unify coarse and fine-grained control signals, allowing flexible and controllable generation of detailed expressions in conjunction with MoEE.
    \item Extensive experiments validate the superiority of our dataset and MoEE model architecture, achieving state-of-the-art emotional expression in audio-driven portrait animation.
\end{itemize}


\section{Related Work}



\begin{figure*}[!ht]  
    \centering
    \includegraphics[width=\linewidth]{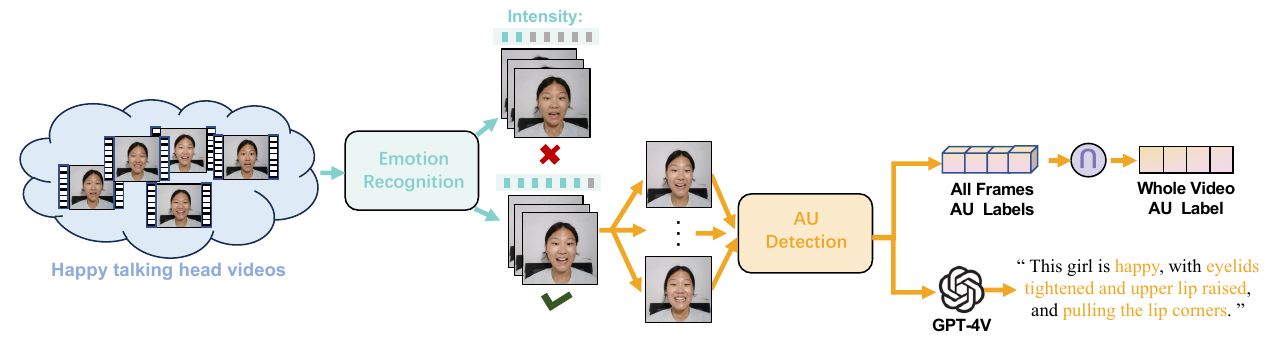}  
    \caption{First, we filter the emotion dataset based on emotion intensity. Then, to achieve fine-grained control, we extract the AUs and prompt GPT-4V to paraphrase them into a sentence.}
    \label{fig:dataset_filter}
    \vspace{-6mm}

\end{figure*}

\textbf{Audio-driven Talking Head Generation}
Significant progress has been made in audio-driven talking head generation and portrait image animation, emphasizing realism and synchronization with audio inputs \cite{chen2024echomimic, sun2023vividtalk, tian2024emo, shen2023difftalk, stypulkowski2024diffused, he2023gaia, corona2024vlogger}. Notably, Wav2Lip \cite{prajwal2020lip} excels at overlaying synthesized lip movements onto existing video content, though its outputs can sometimes show realism issues, such as blurring or distortion in the lower facial region. Recent advancements have pivoted towards image-based methodologies, undergirded by diffusion models, to mitigate the limitations of video-centric approaches and enhance the realism and expressiveness of synthesized animations. Subsequent methods like SadTalker \cite{zhang2023sadtalker} and VividTalk \cite{sun2023vividtalk} incorporated 3D motion modeling and head pose generation to enhance expressiveness and temporal synchronization. AniPortrait \cite{wei2024aniportrait}, EchoMimic \cite{chen2024echomimic}, Hallo \cite{xu2024hallo} and Loopy \cite{jiang2024loopy} have contributed to enhanced capabilities, focusing on expressiveness, realism, and identity preservation.

\textbf{Emotion Editing in Talking Head Videos}
Acknowledging the limitations of previous efforts, which often produce emotionless avatars, there is a growing interest in integrating emotions into talking face generation \cite{peng2023emotalk, liang2022expressive, sun2024fg, yin2022styleheat}. For example, EAT \cite{gan2023efficient} employs a mapping network to derive emotional guidance from latent codes, where EAMM \cite{ji2022eamm} characterizes the facial dynamics of reference emotional videos as displacements in motion representations. Similarly, StyleTalk \cite{ma2023styletalk} develops a style encoder to capture the style of a reference video. However, these methods either support a limited range of individual emotion types or require users to find another video with the desired style, thereby restricting the flexibility and controllability of the generated avatars. Recently, some efforts have focused on using text as an emotion control signal, such as InstructAvatar \cite{wang2024instructavatar}. Nevertheless, these methods rely solely on text descriptions for emotion control, which limits the diversity of control conditions. In contrast, our approach combines advancements in model architecture and dataset design to enable vivid and natural emotion and expression control under multimodal conditions, achieving greater flexibility and expressiveness.


\section{Data Collection and Filteration}


Talking head generation model is naturally scalable for large datasets, but it also requires high-quality data as it learns from data distribution. We have assembled our dataset from a variety of sources. Some open-source emotion datasets, like MEAD \cite{wang2020mead}, have limitations, such as small dataset sizes and a restricted range of emotion categories. To fill the gaps in existing datasets, we have additionally collected a new emotional dataset: DH-FaceEmoVid-150, which focuses on basic emotions and compound emotions. As shown in the Figure~\ref{fig:dataset1}, the entire dataset comprises six basic emotions: angry, disgusted, fear, happy, sad, surprised. In addition, the dataset includes four compound emotions: angrily disgusted, sadly surprised, sadly fear, happily surprised. The total duration of the dataset is 150 hours, featuring 80 actors, with a resolution of no less than 1080×1080. The overview of all the dataset statistics this model used is demonstrated in Tab.~\ref{Table:dataset}. Additional visualizations of the dataset can be found in Appendix~\ref{app:Dataset}

On this basis, to improve the quality of the dataset, we filtered the dataset based on emotion intensity, extracted the corresponding AU (Action Unit) labels for each video, and generated fine-grained text instructions for each video. As shown in Figure~\ref{fig:dataset_filter}, we used LibreFace \cite{chang2024libreface} for emotion recognition to filter out videos with ambiguous emotions. We found that existing emotional talking head datasets typically provide only tag-level annotations with limited emotion categories (\eg, “happy” or “sad”) for talking videos, so we obtained fine-grained AU labels and natural text instructions through Action Unit Extraction and VLM Paraphrase. Action Units (AUs) are defined by Facial Action Coding System (FACS) \cite{ekman1978facial} and are used to describe facial muscle movements. We extracted AU labels for each frame and the corresponding action descriptions using the ME-GraphAU model \cite{luo2022learning}. Then, we randomly selected 5 frames from a video, input the action descriptions and frames images into the GPT-4V \cite{gpt4v}, and obtained text instructions for the entire video.

\begin{table}[t]
\centering
\vspace{-3mm}
\resizebox{0.48\textwidth}{!}{
\begin{tabular}{lcccccc}
\toprule
Datasets    & IDs & Hours & Single Emotion & Compound Emotion & Text & AU Label \\ \midrule
HDTF~\cite{zhang2021flow}        & 362 & 16    & $\times$      & $\times$       & $\times$     & $\times$  \\ 
DH-FaceVid-1K~\cite{di2024facevid1klargescalehighqualitymultiracial}   & 500 & 200   & $\times$     & $\times$        & $\times$        & $\times$     \\ 
Mead~\cite{wang2020mead}   & 60 & 40   & \checkmark       & $\times$          & $\times$     & $\times$  \\ 
DH-FaceEmoVid-150 & 80 & 150   & \checkmark    & \checkmark          & \checkmark     & \checkmark\\ \bottomrule
\end{tabular}
}
\caption{Statistics of the collected dataset}
\label{Table:dataset}
\vspace{-5mm}
\end{table}

\section{Methodology}
\begin{figure*}[t]  
    \centering
    \includegraphics[width=1.0\linewidth]{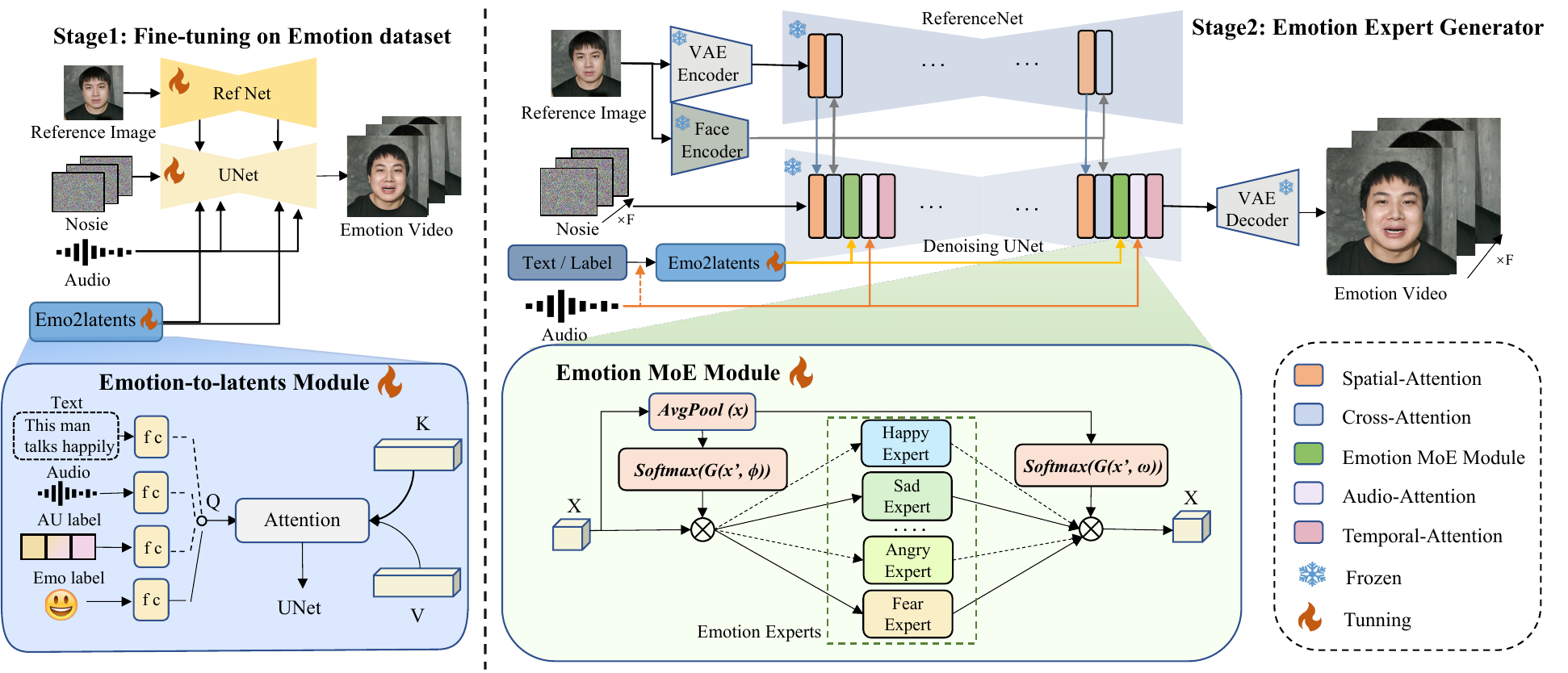}
    \caption{The overall framework of MoEE implements a two-stage training process. First, we fine-tune the Reference Net and denoising U-Net modules on emotion datasets to enable the model to learn as much prior knowledge about expressive faces as possible. Then, we achieve more natural and accurate emotion and expression control through the Mixture of Emotion Experts. Additionally, the Emotion-to-Latents module enables multi-modal emotion control.}
    \label{fig:pipeline}
    \vspace{-5mm}
\end{figure*}
In this section, we introduce our MoEE model. In Sec.~\ref{app:Model Overview}, we provide an overview of the architecture of MoEE. Subsequently, Sec.~\ref{app:Mixture of Emotion Experts} describes the Mixture of Emotion Expert to achieve diverse emotions control, while Sec.~\ref{app:Emotion-to-latents Module} introduce the detail of Emotion-to-latents Module. Finally, in Sec.~\ref{app:Training and Inference}, we demonstrate the training and inference details. 
\subsection{Model Overview}\label{app:Model Overview}

Given one portrait image $\boldsymbol{I}$, a sequence of audio clips $\boldsymbol{A} = [a_1, a_2, ..., a_N ]$, and the Emotion Condition $\boldsymbol{C}$, our model is tasked with animating the portrait to utter the audio with the target style represented by the text, audio or label. Overall, we aim to learn a mapping to generate a video $\boldsymbol{V = F(I,A,C)}$. 
As illustrated in Figure~\ref{fig:pipeline}, the backbone of MoEE is a denoising U-Net \cite{blattmann2023stable}, which denoises the input multi-frame noisy latent under the conditions. This architecture includes an additional Reference Net module, which is a Unet-based Stable Diffusion network with the same number of layers as the denoising model network, to capture the visual appearance of both the portrait and the associated background. The input audio embedding is derived from a 12-layer wav2vec \cite{schneider2019wav2vec} network and fed into the Audio Attention Layer to encode the relationship with the audio. The Temporal Attention layer \cite{guo2023animatediff} is a temporal-wise self-attention layer that captures the temporal relationships between the video frames. Additionally, the denoising U-Net incorporates an Emotion MoE Module, consisting of six emotion-specific expert modules, to encode relationships with specific emotional conditions.
In the following sections, we will detail how to design a mixture of emotion experts module, as well as how to design a multimodal emotion condition module. 

\subsection{Mixture of Emotion Experts}\label{app:Mixture of Emotion Experts}
To address the factors contributing to the poor performance of emotion generation, we propose constructing a Mixture of Experts (MoE) module. This module consists of emotion experts, each specializing in a single emotion. We focus on six basic emotions: happiness, sadness, anger, disgust, fear, and surprise, training an expert model for each. This is achieved by training six cross-attention modules on well-structured, single-emotion datasets. During the training of each expert, the latent noise bypasses the other modules, whose parameters remain fixed. The cross attention operation is formulated as:
\begin{equation}
    CrossAttn(Z_t,C) = \operatorname{softmax}(QK^{T}/{\sqrt{d}})V
\end{equation}
where $Q=W_QZ_t$, $K=W_KC$ and $V=W_VC$ are the queries; $W_Q$, $W_K$ and $W_V$ are learnable projection matrices; and $d_k$ is the dimensionality of the keys.

\noindent\textbf{Soft Mixture Assignment}
After training expert models for the six basic emotions, an assignment mechanism needs to be introduced to better coordinate each expert model during the generation process. Since the process of generating complex emotions requires combinations of multiple basic emotions, while the hard assignment mode in MoE only permits one expert to access the given input. Therefore, inspired by Soft MoE \cite{puigcerver2023sparse}, we adopt a soft assignment, allowing multiple experts to handle input simultaneously. In addition to the need for emotion control over the global video frames, localized control is also essential, as different emotions are mainly expressed through specific facial expressions. Therefore, we use both global and local assignments. Local assignment, by independently assigning weights to each token, enables more precise control over the emotional expression of local features, facilitating the generation of more vivid expressions.

Specifically, considering the input $X \in \mathbb{R}^{b \times n \times d}$ where $b$ is the number of batch size, $n$ is the number of token and $d$ is the feature dimension. For local assignment, we use a local gating network that contains a learnable gating layer $G(\,, \phi)$ ($\phi \in \mathbb{R}^{d \times e}$, $e$ is the number of experts and here $e$ is $6$, below is the same) and a $\operatorname{sigmoid}$ function. The gating network is to  produce six normalized score maps $s = [s_1, s_2, s_3, s_4, s_5, s_6]$ ($s \in \mathbb{R}^{n \times e}$) for six emotion experts as formulated:
\begin{equation}
    s = \operatorname{sigmoid}(G(X \,, \phi)
\end{equation}
For global soft assignment, we use a global gating network including an AdaptiveAvePool, a learnable gating layer $G(\,, \omega)$ ($\omega \in \mathbb{R}^{d \times e}$), and a $\operatorname{softmax}$ function. This gating network is to produce six global scalars $g=[g_1, \; g_2, \; g_3, \; g_4, \; g_5, \; g_6]$($g \in \mathbb{R}^{e}$) for six experts as formulated:
\begin{equation}
    g = \operatorname{softmax}(G(\operatorname{Pool}(X) \,, \omega)) \,.
\end{equation}
The soft mechanism is built on the fact that the input $X$ can adaptively determine how much (weight) should be sent to each expert by the $\operatorname{softmax}$ function. At the same time, the weights among the various emotion experts are interdependent. For single emotion control generation, only the corresponding expert model needs to be invoked. For compound emotion control generation, cooperation among different emotion expert modules is required. Specifically, we first send $X$ to each emotion expert respectively, use $s_i$, where $i$ is the number of emotion expert, to perform element-wise multiplication (local assignment), and also perform global control by scalars $g_i$. Then we add them back to the $X$, producing a new output $X^{'}$:
\begin{equation} \label{final output}
    X^{'} = X + \sum_{i=1} g_i \cdot E_i(X \cdot s_i) \quad
\end{equation}
Futher network details about Mixture of Emotion Experts are provided in the Appendix \ref{app:network}

\subsection{Emotion-to-latents Module}\label{app:Emotion-to-latents Module}
This module can accept a variety of different modality conditions, such as text, audio, and labels. As illustrated in Figure~\ref{fig:pipeline}, it maps multimodal conditions, which have both strong and weak correlations with emotion, to a shared emotion latent space, ensuring that all these conditions have a good control effect. Specifically, the conditions from various modalities are first transformed into feature vectors using pretrained encoders. For textual information, we use the T5 text encoder \cite{ni2021sentence} and for audio signals, we employ emotion2vec model \cite{ma2023emotion2vec}. For label signals, we train a custom two-layer MLP network as an encoder. Then, we map these feature vectors to the same dimension via fully connected layers (FC) to serve as query latent in the attention mechanism. Additionally, we maintain a set of learnable embeddings, which act as key and value features for attention computation to obtain new features. These new features, referred to as emotion latent, replace the input conditions in subsequent computations. These feature vectors are then fed into the Emotion MoE Module, serving as key and value features, injecting condition information into the U-Net. Futher network details about Emotion-to-latents Module are provided in the Appendix \ref{app:network}

\subsection{Training and Inference}\label{app:Training and Inference}

\begin{figure}[t]  
\centering
\includegraphics[width=1.0\linewidth]{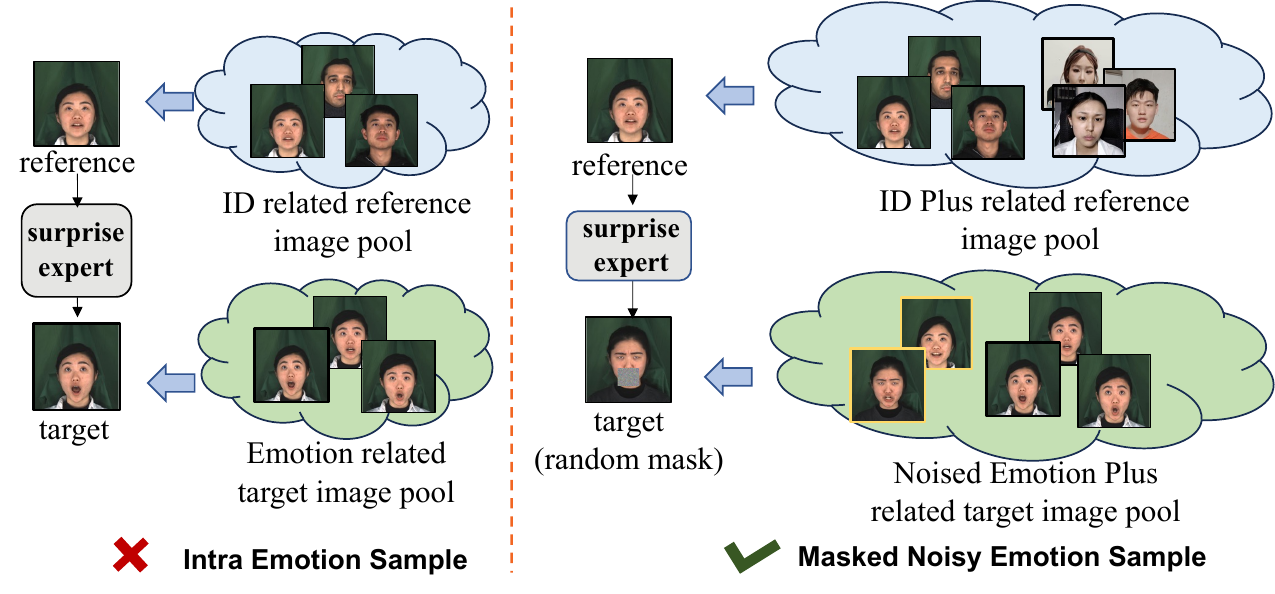}  
\caption{Visualization of different emotion sample strategy. Results demonstrate that the proposed masked noisy emotion sample strategy can ensure natural and vivid expression.
}
    \label{fig:sample}
    \vspace{-3mm}
\end{figure}

\noindent\textbf{Masked Noisy Emotion Sampling}
Due to the relatively small size of the segmented single-emotion datasets, the expert models trained on them often risk learning knowledge beyond the targeted emotion. To amplify the impact of emotion, we incorporate data of other emotions or neutral expressions with a certain probability for a particular emotion. As illustrated in Figure~\ref{fig:pipeline}, this approach introduces noise into the emotion conditions while expanding the number of person IDs thus enabling the model to concentrate more effectively on the control of emotion conditions. 

However, since frames with different emotions come from different speech videos, there can be significant variations in mouth shapes, leading the model to focus more on mouth transformations rather than the emotions themselves. To mitigate this, we first locate the position of the mouth in portrait images by Mediapipe \cite{lugaresi2019mediapipe}. We then apply a mask to the mouth. This method ensures that the expert model maintains consistent person ID while achieving more accurate learning of facial expression transformations.


\noindent\textbf{Loss function}
Since the resolution of latent space ($64 \times 64$ for $512 \times 512$ image) is relative too low to capture the subtle facial details, a timestep-aware spatial loss is proposed to learning the face expression directly in the pixel space. The loss function $L$ during the training process is summarized as:
\begin{equation}
L = L_{latent} + \lambda L_{spatial}
\end{equation}
Where $L_{latent}$ is the objective function guiding the denoising process. With timestep $t$ is uniformly sampled from $\left \{ 1, ..., T  \right \} $. The objective is to minimize the error between the true noise \(\epsilon\) and the model-predicted noise \(\epsilon_{\theta}(z_{t}, t, c)\) based on the given timestep \(t\), the noisy latent variable \(z_t\), and $c_{embed}$ denotes the conditional embeddings. The loss function during the training process is summarized as:
\begin{equation}
L_{latent} = \mathbb{E}_{t,c_{embed},z_t,\epsilon}[||\epsilon-\epsilon_{\theta}(z_t,t,c_{embed})||^{2}]
\end{equation}
For $L_{spatial}$, predicted latent $z_t$ is first mapped to $z_0$ by sampler and the predicted image $I_{p}$ is obtained via passing $z_0$ to the vae decoder. The L1 loss is computed on the predicted image and its corresponding ground truth. Additionally, we introduce a Perception loss, enhancing details and visual realism. The specific formula is shown below:
\begin{equation}
L_{spatial} = w(t) (||I_{p}, I_{GT}|| + ||V(I_{p}), V(I_{GT})||^{2})
\end{equation}
where $V$ denotes the feature extractor of VGG19, and $w(t) = cosine(t* \pi / 2T)$ serves as a timestep aware function to reduce the weight for large $t$.

\begin{figure*}[t]  
    \centering
    \includegraphics[width=1.0\linewidth]{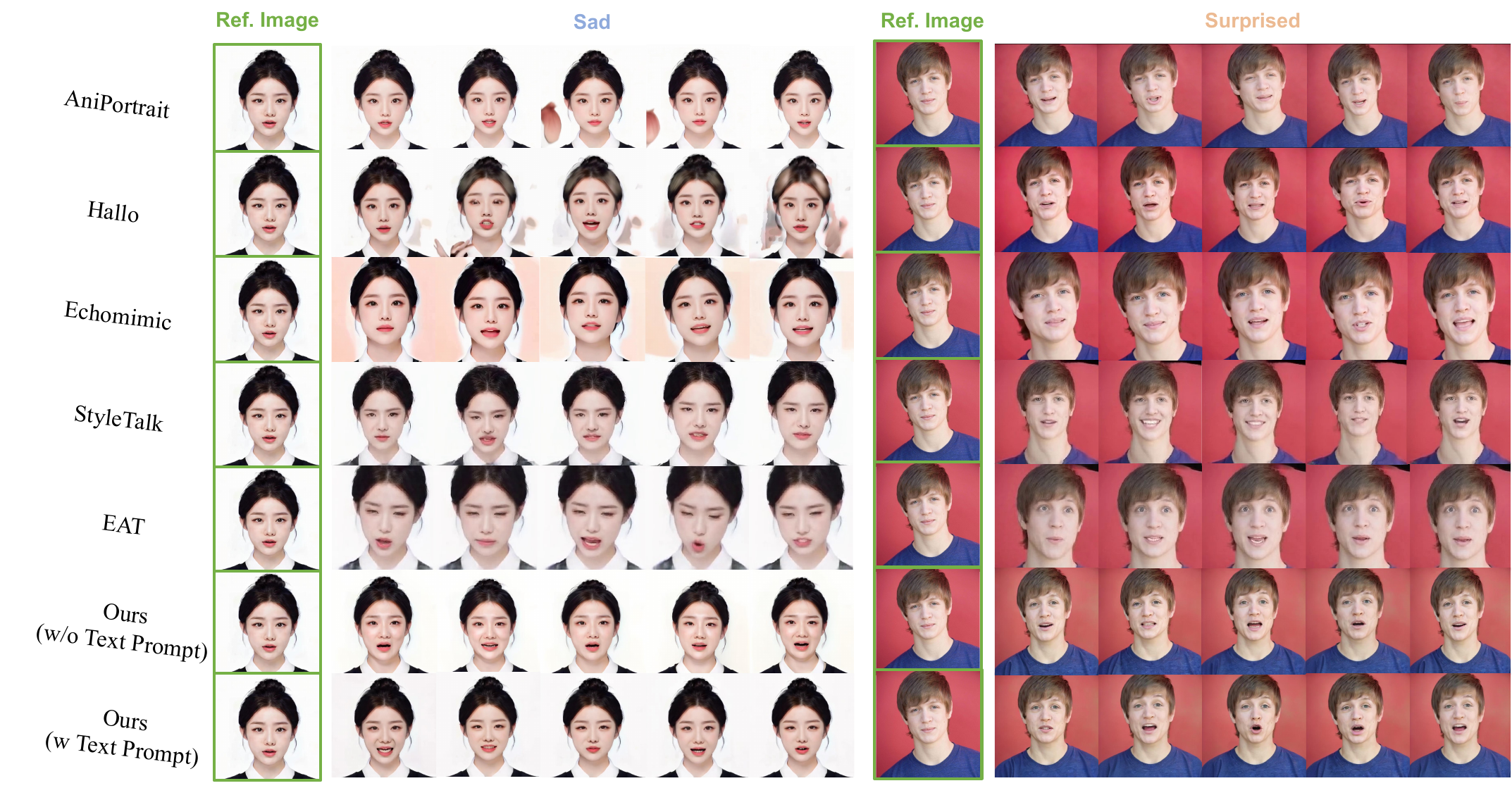}  
    \caption{Qualitative comparison with baselines. It shows that MoEE achieves well lip-sync quality and emotion controllability. Additionally, MoEE enables emotion control without text prompt.}
    \label{fig:main_qualtive_img}
    \vspace{-5mm}

\end{figure*}

\noindent\textbf{Training}
This study implements a two-stage training process aimed at optimizing distinct components of the overall framework. Next, We will describe each stage in detail below.

In the initial stage, considering the limited emotion prior for diffusion model could be caused by the absence of high-quality emotion datasets, our work bridges this gap by providing carefully collected emotion datasets. We fine-tuning the Reference Net and denoising U-Net modules on these emotion Datasets, and the well-trained model is then sent to the next stage.

In the second stage, the model is trained to generate video frames based on the reference image, driven audio, and emotion condition. During this phase, the spatial modules, cross modules, audio modules and temporal modules of Reference Net and denoising U-Net remain static. The optimization process focuses on the Emotion MoE modules, which are optimized to enhance the accuracy and diversity of emotion control generation capabilities.

\noindent\textbf{Inference}
During the inference stage, the network takes a single reference image, driving audio and emotion condition as input, producing a video sequence that animates the reference image based on the corresponding emotion and audio. To produce visually consistent long videos, we utilize the last 2 frames of the previous video clip as the initial k frames of the next clip, enabling incremental inference for video clip generation.

\section{Experiment}
\subsection{Experiment setup}

\noindent\textbf{Datasets} As outlined earlier, this research utilizes the filtered HDTF \cite{zhang2021flow}, a segment of the DH-FaceVid-1K \cite{di2024facevid1klargescalehighqualitymultiracial} dataset, as well as MEAD \cite{wang2020mead} and DH-FaceEmoVid-150 for training. During the testing phase, we choose 20 subjects from the HDTF datasets, selecting 10 video clips for each subject at random, with each clip lasting about 5 to 10 seconds. Notably, to assess the efficacy of the mixture of emotion experts, we extract 20 subjects from both MEAD and DH-FaceEmoVid-150 datasets. For each of these subjects, 10 videos are randomly chosen from 4 out of the 6 available emotional categories.

\noindent\textbf{Evaluation Metrics} The evaluation metrics utilized in the portrait image animation approach include Learned Perceptual Image Patch Similarity (LPIPS) \cite{zhang2018lpips}, Fréchet Inception Distance (FID) \cite{FIDheusel2017gans}, Fréchet Video Distance (FVD) \cite{wang2018videotovideosynthesis} , Average Keypoint Distance (AKD) \cite{AKDsiarohin2019animating}
, and synchronization metrics Sync-C and Sync-D \cite{sync}. LPIPS measures perceptual similarity, with lower scores being preferable. FID and FVD evaluate realism, where lower scores indicate greater similarity to real data. AKD measures the alignment of facial keypoints, with lower values indicating higher accuracy in reproducing facial expressions and movements. Sync-C and Sync-D evaluate lip synchronization, where higher Sync-C and lower Sync-D scores reflect better audio-visual alignment. Since the dataset’s audio is in Chinese, Sync-C and Sync-D may not provide an entirely objective evaluation, so we used tailored combinations of evaluation metrics for each dataset.

\noindent\textbf{Baseline} We compared our proposed method with publicly available implementations of AniPortrait \cite{wei2024aniportrait}, Hallo \cite{xu2024hallo}, EchoMimic \cite{chen2024echomimic}, StyleTalk \cite{ma2023styletalk}, EAT \cite{gan2023efficient} on the HDTF, MEAD, and DH-FaceEmoVid-150 datasets. We also conducted a qualitative comparison to provide deeper insights into our method's performance and its ability to generate realistic, expressive talking head animations.

More experiment details can be found in the Appendix.



\subsection{Quantitative Results}
We conducted an exhaustive comparative analysis of existing diffusion-based methods driven by audio. This study underscored the efficacy of MoEE in achieving vivid and accurate control over emotions and expressions. To ensure rigorous evaluation across datasets, we selected appropriate metrics to showcase each method’s unique strengths.

\begin{table}[t]
\centering
\resizebox{0.48\textwidth}{!}{
\begin{tabular}{lccccccc}
\toprule
\multirow{2}{*}{\textbf{Method}} & \multicolumn{5}{c}{\textbf{HDTF}}                                                \\ \cline{2-6} 
            & \textbf{FID$\downarrow$} & \textbf{FVD$\downarrow$} & \textbf{LPIPS$\downarrow$} & \textbf{Sync-C$\uparrow$} & \textbf{Sync-D$\downarrow$} \\ \hline
AniPortrait \cite{wei2024aniportrait}               & 36.826            & 476.818            & 0.211          & 5.977           & 9.898               \\
Hallo \cite{xu2024hallo}                            & \textbf{28.605}            & 343.023            & 0.167          & \textbf{6.254}           & \textbf{8.735}               \\
Echomimic \cite{chen2024echomimic}                  & 48.323            & 526.125            & 0.312          & 6.73            & 9.157               \\
StyleTalk \cite{ma2023styletalk}                    & 76.564            & 502.334            & 0.326          & 3.885           & 10.644               \\
EAT \cite{gan2023efficient}                         & 81.254            & 545.266            & 0.357          & 5.012           & 9.885               \\
MoEE (Ours)                                               & 28.834            & \textbf{322.625}            & \textbf{0.152}          & 6.114           & 9.101               \\ \bottomrule
\end{tabular}
}
\caption{Comparison of various methods on the HDTF dataset.}
\label{tab:hdtf}
\vspace{-2mm}
\end{table}

\noindent\textbf{Comparison on the HDTF dataset} Table~\ref{tab:hdtf} indicates that MoEE outperforms other methods across multiple metrics. The Sync-C and Sync-D metrics show a slight decline due to the Chinese talking head dataset we used.

\begin{table}[t]
\centering
\resizebox{0.48\textwidth}{!}{
\begin{tabular}{lccccccc}
\toprule
\multirow{2}{*}{\textbf{Method}} & \multicolumn{5}{c}{\textbf{MEAD}}                                                \\ \cline{2-6} 
& \textbf{FID$\downarrow$} & \textbf{FVD$\downarrow$} & \textbf{LPIPS$\downarrow$} & \textbf{Sync-C$\uparrow$} & \textbf{Sync-D$\downarrow$} \\ \hline
AniPortrait \cite{wei2024aniportrait}            & 64.582       & 735.691          & 0.379          & 6.212               & 9.866               \\
Hallo \cite{xu2024hallo}                         & 72.384       & 715.119          & 0.356          & 6.856           & 9.103               \\
Echomimic \cite{chen2024echomimic}               & 75.678       & 699.322          & 0.413          & 6.668               & 9.342               \\
StyleTalk \cite{ma2023styletalk}                 & 49.399       & 577.657          & 0.295          & 3.968           & 10.578          \\
EAT \cite{gan2023efficient}             & \textbf{32.696}       & 412.709          & 0.394          & 5.509           & 8.209               \\
MoEE (Ours)                                            & 39.420  & \textbf{403.416}   & \textbf{0.288}    & \textbf{7.002}               & \textbf{8.105}               \\ \bottomrule
\end{tabular}
}
\caption{Comparison of various methods on the MEAD dataset.}
\label{tab:mead}
\vspace{-2mm}
\end{table}

\noindent\textbf{Comparison on the MEAD dataset}
Table~\ref{tab:mead} indicates that MoEE can achieve the best performance in most metrics, demonstrates superior and stable performance across image and video quality, as well as motion synchronization.

\begin{table}[t]
\centering
\resizebox{0.48\textwidth}{!}{
\begin{tabular}{lccccccc}
\toprule
\multirow{2}{*}{\textbf{Method}} & \multicolumn{4}{c}{\textbf{DH-FaceEmoVid-150}}                                                \\ \cline{2-5} 
            & \textbf{FID$\downarrow$} & \textbf{FVD$\downarrow$} & \textbf{LPIPS$\downarrow$} & \textbf{AKD$\downarrow$} & \\ \hline
AniPortrait \cite{wei2024aniportrait}            & 66.034       & 712.291      & 0.323          & 20.654                             \\
Hallo \cite{xu2024hallo}                         & 72.354       & 702.841      & 0.329          & 16.444                             \\
Echomimic \cite{chen2024echomimic}               & 71.288       & 653.927      & 0.366          & 8.332                             \\
StyleTalk \cite{ma2023styletalk}                 & 51.038       & 534.217      & 0.242          & 4.234                            \\
EAT \cite{gan2023efficient}                      & 48.011       & 467.739      & 0.260          & 14.109                             \\
MoEE (Ours)                                            & \textbf{39.619}       & \textbf{402.803}      & \textbf{0.182}          & \textbf{4.028}                             \\ \bottomrule
\end{tabular}
}
\caption{Comparison of various methods on the DH-FaceEmoVid-150 dataset.}
\label{tab:DH-FaceEmoVid-150}
\vspace{-2mm}
\end{table}

\noindent\textbf{Comparison on the DH-FaceEmoVid-150 dataset} 
DH-FaceEmoVid-150 dataset contains a large set of facial videos with various emotions, enabling MoEE to generate more stable results. As shown in Table~\ref{tab:DH-FaceEmoVid-150}, MoEE significantly outperforms all other methods across all metrics. 

\subsection{Qualitative Results}

\begin{figure}[t]  
\centering
\includegraphics[width=0.9\linewidth]{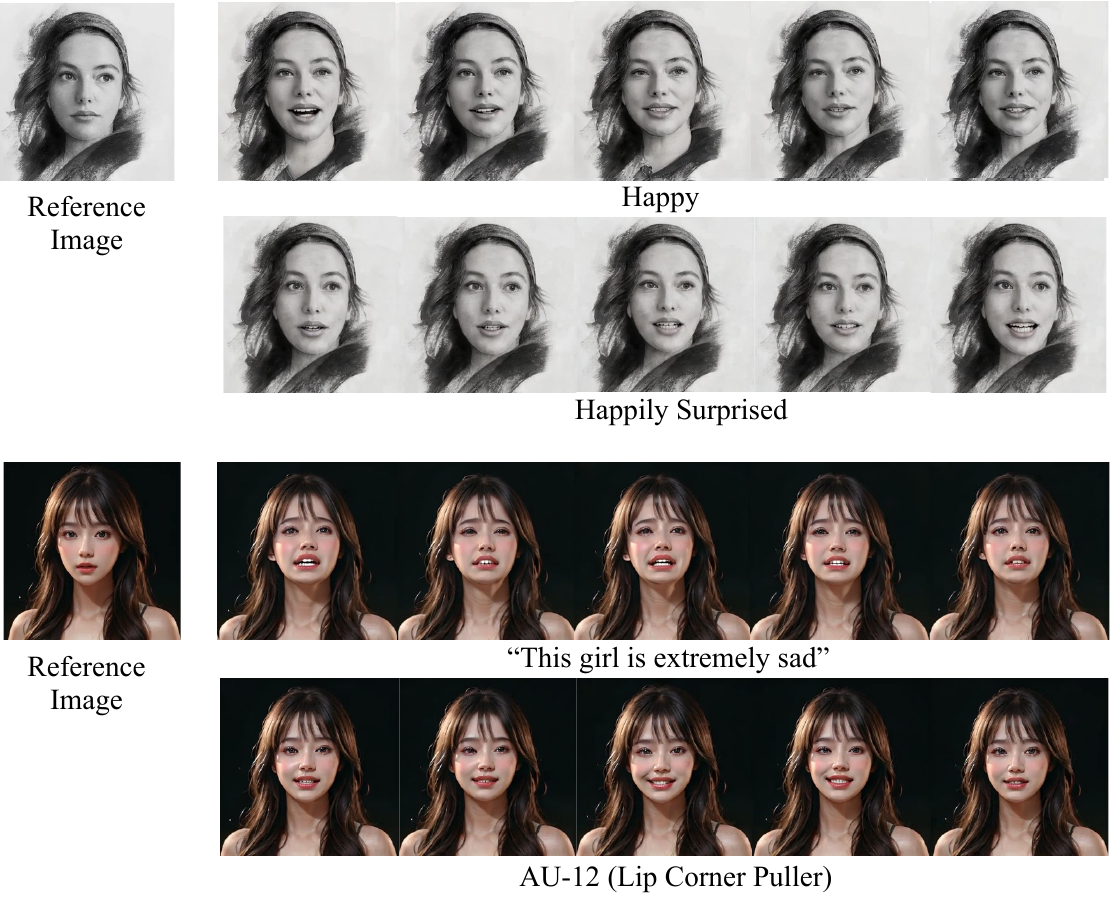}  
\caption{Portrait image animation results on Emotion Control and AU Control with different portrait styles.
}
    \label{fig:style_img}
    \vspace{-2mm}
\end{figure}

\noindent\textbf{Visual comparison with baselines}
Figure~\ref{fig:main_qualtive_img} presents a visual comparison of MoEE against other methods across different emotion control, demonstrating superior performance under different conditions. Notably, AniPortrait, Hallo and EchoMimic often exhibit background noise issues, while EAT and StyleTalk often produce incorrect and unnatural expressions. Our model demonstrates strong lip-sync accuracy, along with enhanced naturalness and clarity, while effectively preserving identity characteristics and background stability.

\noindent\textbf{Visualization on Emotion Control and AU Control}
Figure~\ref{fig:style_img} illustrates that our method can achieve single-emotion control, compound emotion control, and fine-grained expression control based on AU labels. Additionally, our method is capable of processing a wide range of input types, including paintings, portraits from generative models and more. These findings highlight the versatility and effectiveness of our approach in emotion control and accommodating different artistic styles. More experiment details can be found in the Appendix~\ref{app:Visualization}

\begin{table}[t]
\vspace{-0mm}
\begin{center}
\resizebox{0.48\textwidth}{!}{
\vspace{-0.0mm}
	\begin{tabular}{lcccc}
		\toprule
		 \textbf{Methods}    & {\textbf{FID}$\downarrow$} &  {\textbf{FVD}$\downarrow$} & {\textbf{LPIPS}$\downarrow$} & {\textbf{AKD}$\downarrow$}  \\  \midrule
        
        (a) w/o MoEE         & 58.411      & 655.329      & 0.325      & 14.959       \\
        (b) w/o GS Assigment             & 46.334      & 447.814     & 0.194      & 10.652       \\
        (c) w/o MNS Sample        & 51.788      & 489.241      & 0.211      & 4.591       \\
        (d) w/o DH-FaceEmoVid-150        & 52.412      & 511.928      & 0.275      & 7.885       \\
        
        \midrule
        ~MoEE (Ours)                       & \textbf{39.619} & \textbf{402.803} & \textbf{0.182} & \textbf{4.028} \\
             
		\bottomrule
	\end{tabular}
}
\caption{More ablation studies on the proposed techniques. The first row shows the results without using Mixture of Emotion Experts (MoEE). The second row presents the results without Global Soft Assignment (GS Assignment). The third row displays the results without Masked Noisy Emotion Sample (MNS Sample). The forth row displays the results without DH-FaceEmoVid-150 dataset.}
\label{tab:Ablation Results_2}
\end{center}
\vspace{-1mm}
\end{table}

\subsection{Ablation Study}

\begin{figure}[t]  
\centering
\includegraphics[width=0.9\linewidth]{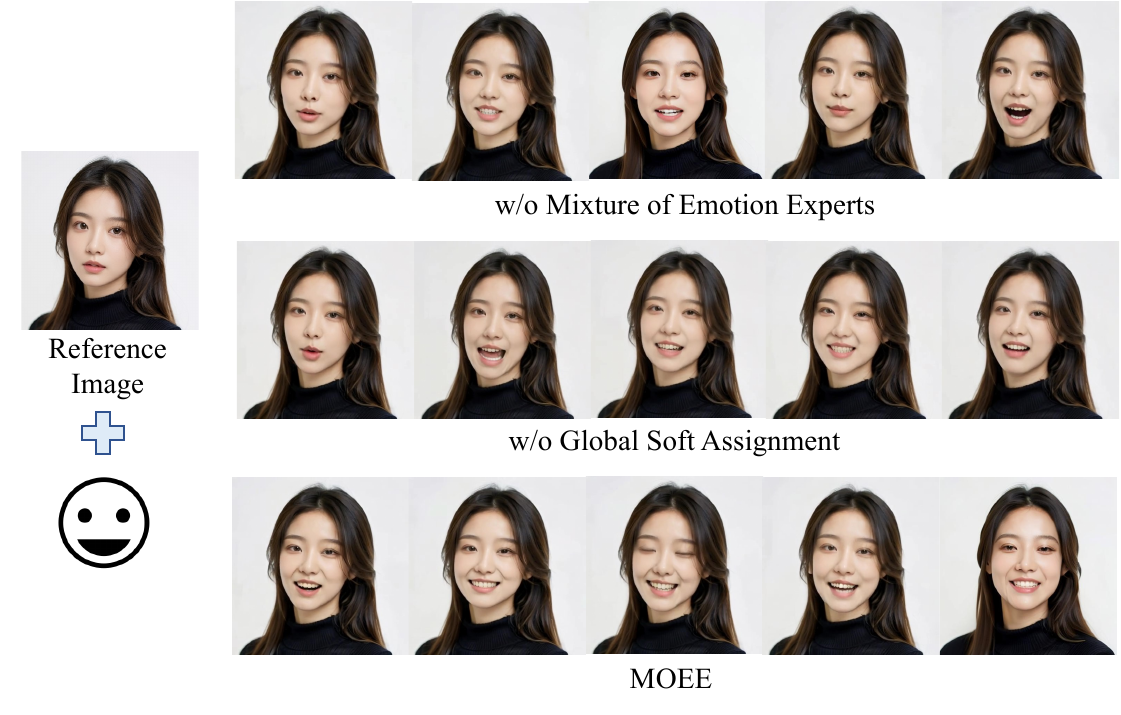}  
\caption{Visualization of different emotion control methods. Results demonstrate that the proposed mixture of experts methods can ensure natural and vivid expression.
}
    \label{fig:abaltionMOE}
    \vspace{-2mm}
\end{figure}

\begin{figure}[t]  
\centering
\includegraphics[width=1\linewidth]{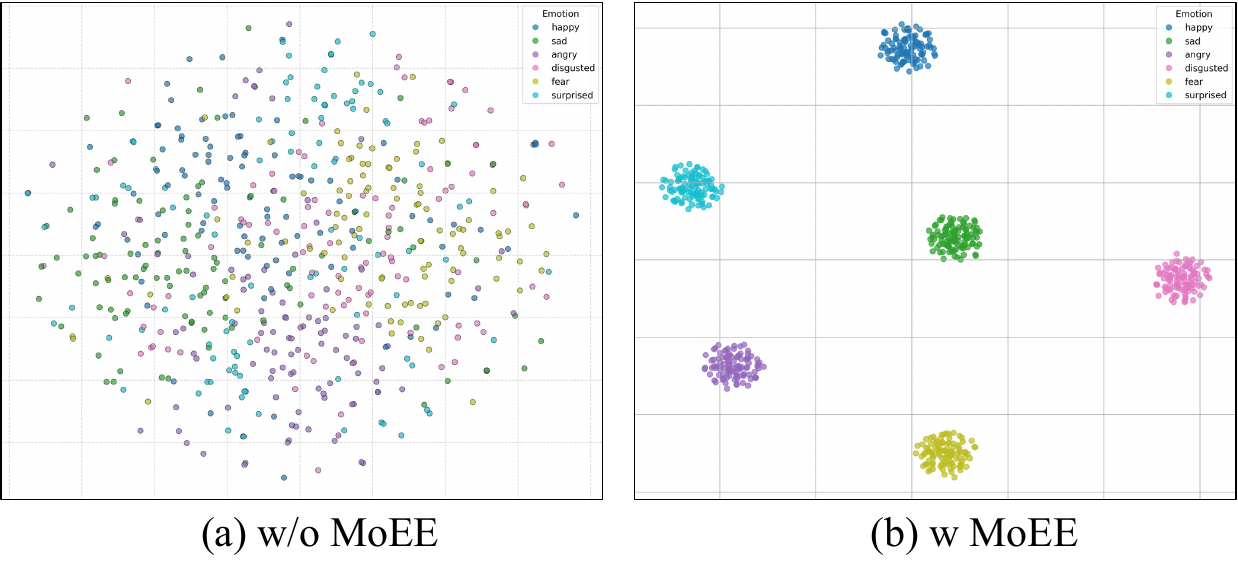}  
\caption{Visualization of the latent space.}
    \label{fig:latents_space}
    \vspace{-2mm}
\end{figure}

\noindent\textbf{Mixture of Emotion Experts Method} 
We analyze the impact of different emotion control methods on improving emotion control. As shown in Figure~\ref{fig:abaltionMOE}, we compare three approaches: without mixture of experts models, without global assignment, and with both mixture of experts models and global assignment. The experiments demonstrate that our method effectively capture fine-grained expression details. Table~\ref{tab:Ablation Results_2} illustrates the effects of Mixture of Emotion Experts. Notably, using Global Soft Assignment enhances visual quality and emotion control, as reflected by improvements in FID, FVD, and AKD.

We also analyzed the effect of employing the Mixture of Emotion Experts module on the latent space across different emotional conditions. Figure~\ref{fig:latents_space} presents a visualization of the latent space: with the application of this module, the latent distributions across different emotional conditions exhibit greater separation, ensuring accurate and stable generation of talking heads with specific emotions.


\begin{figure}[t]  
\centering
\includegraphics[width=1\linewidth]{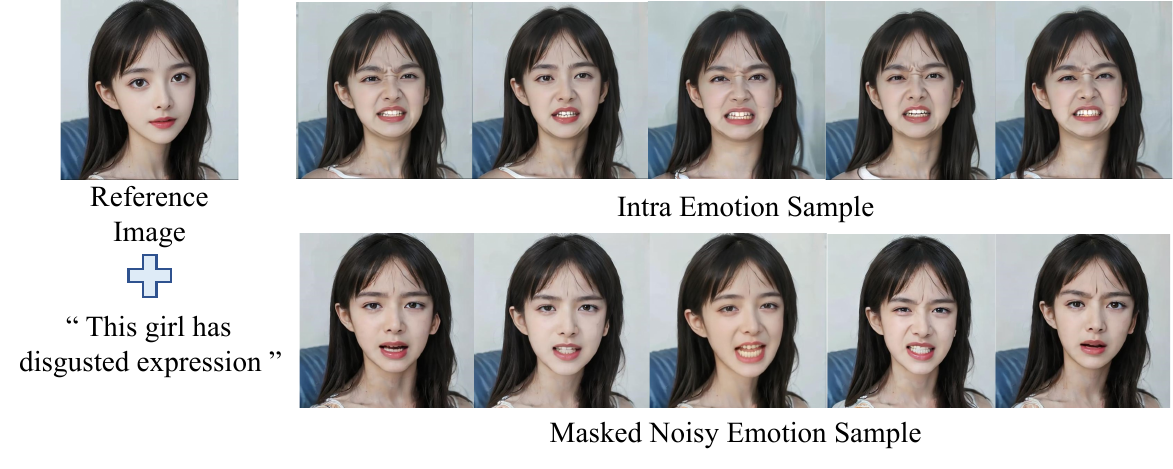}  
\caption{Visualization of different emotion sample strategy. Results demonstrate that the proposed masked noisy emotion sample strategy can ensure natural and vivid expression.
}
    \label{fig:ablation_maskSample}
    \vspace{-2mm}
\end{figure}

\noindent\textbf{Masked Noisy Emotion Sample Strategy} 
We introduced a masked noisy emotion sample strategy to enhance emotion accuracy and character ID consistency. As illustrated in Figure~\ref{fig:ablation_maskSample}, we compared two sampling methods: (a) intra-emotion sampling and (b) masked noisy emotion sampling and the experimental results demonstrate that our proposed approach effectively generates accurate and natural expressions. Table~\ref{tab:Ablation Results_2} further validates the effectiveness of the masked noisy emotion sampling strategy.

\begin{figure}[t]  
\centering
\includegraphics[width=1\linewidth]{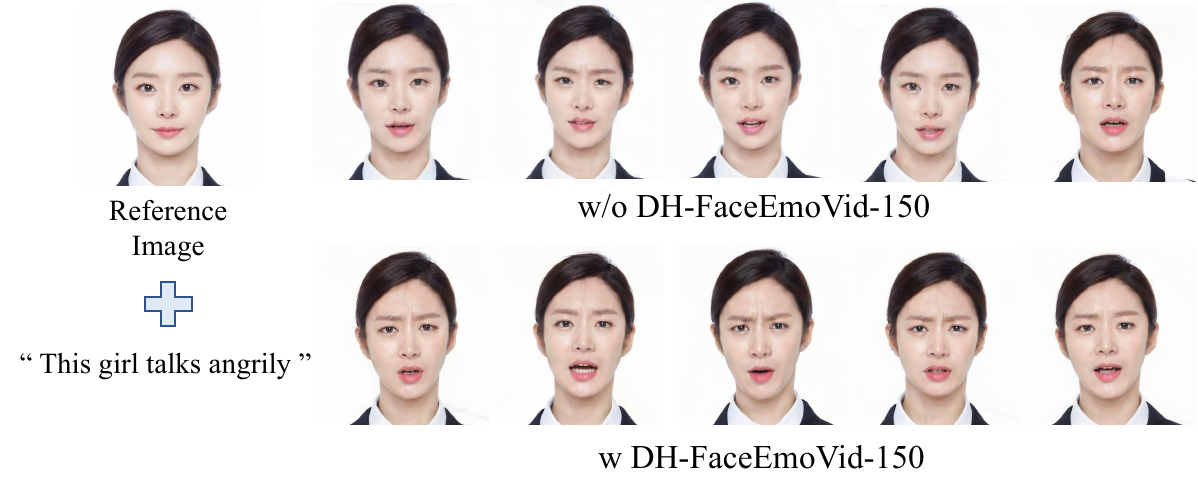}  
\caption{Visualization of emotion control based on different datasets. Results demonstrate that DH-FaceEmoVid-150 can enhance emotion control ability.}
    \label{fig:ablationdataset}
    \vspace{-2mm}
\end{figure}

\noindent\textbf{Dataset Efficiency}
To evaluate the effectiveness of the dataset, we conducted both quantitative and qualitative comparison experiments. As illustrated in Figure~\ref{fig:ablationdataset} and Table~\ref{tab:Ablation Results_2}, the results verified that DH-FaceEmoVid-150 significantly enhances the ability of emotional Audio-Driven Portrait Animation.

\subsection{User Study}
We conduct a user study to compare our method with the other methods. We involve 15 experienced users to score the generation quality and controllability of each model. We randomly select 10 generation examples from the MEAD and DH-FaceEmoVid-150 datasets. Our evaluation metric is the Mean Opinion Score (MOS). We assess the lip-sync quality (Lip.), emotion controllability (Emo.), naturalness (Nat.) and identity preservation (ID.). Participants were presented with one video at a time and asked to rate each video for each score on a scale of 1 to 5. We calculated the average score as the final result. As shown in Table~\ref{tab:user_study_mead} and Table~\ref{tab:user_study_facevid}, Our proposed method achieves the best results across all evaluation criteria.

\begin{table}[t]
\vspace{-0mm}
\begin{center}
\vspace{-0.0mm}
\resizebox{0.48\textwidth}{!}{
	\begin{tabular}{lcccc}
		\toprule
		 \textbf{Method}    & {\textbf{Emo.}$\uparrow$} &  {\textbf{Lip.}$\uparrow$} & {\textbf{Nat.}$\uparrow$} & {\textbf{ID.}$\uparrow$}  \\  \midrule
        
        AniPortrait~\cite{wei2024aniportrait}               & 2.05      & 3.98      & 4.33      & 4.31       \\
        Hallo~\cite{xu2024hallo}             & 2.03      & 4.48      & 4.25      & 4.05       \\
        Echomimic~\cite{chen2024echomimic}         & 2.88      & 4.71      & 4.65      & 4.25       \\
        StyleTalk~\cite{ma2023styletalk}            & 4.44      & 3.22      & 3.54      & 3.58       \\
        EAT~\cite{gan2023efficient}                              & 4.23      & 3.57      & 3.81      & 4.29       \\
        
        \midrule
        MoEE (Ours)                       & \textbf{4.65} & \textbf{4.74} & \textbf{4.71} & \textbf{4.55} \\
             
		\bottomrule
	\end{tabular}
    }
\end{center}
\caption{User Study for MoEE and other baselines on MEAD. The bold values indicate the best results. }
\label{tab:user_study_mead}
\vspace{-1mm}
\end{table}

\begin{table}[t]
\vspace{-0mm}
\begin{center}
\vspace{-0.0mm}
\resizebox{0.48\textwidth}{!}{
	\begin{tabular}{lcccc}
		\toprule
		 \textbf{Method}    & {\textbf{Emo.}$\uparrow$} &  {\textbf{Lip.}$\uparrow$} & {\textbf{Nat.}$\uparrow$} & {\textbf{ID.}$\uparrow$}  \\  \midrule
        
        AniPortrait~\cite{wei2024aniportrait}               & 2.25      & 4.02      & 4.18      & 4.31       \\
        Hallo~\cite{xu2024hallo}             & 2.13      & 4.21      & 4.54      & 4.33       \\
        Echomimic~\cite{chen2024echomimic}         & 2.78      & 4.56      & 4.71      & 4.25       \\
        StyleTalk~\cite{ma2023styletalk}            & 4.62      & 3.51      & 3.73      & 3.58       \\
        EAT~\cite{gan2023efficient}                              & 4.51      & 3.45      & 3.75      & 4.21       \\
        
        \midrule
        MoEE (Ours)                       & \textbf{4.73} & \textbf{4.81} & \textbf{4.76} & \textbf{4.62}  \\
             
		\bottomrule
	\end{tabular}
    }
\end{center}
\caption{User Study for MoEE and other baselines on DH-FaceEmoVid-150. The bold values indicate the best results. }
\label{tab:user_study_facevid}
\vspace{-1mm}
\end{table}

\section{Limitations}

Our method still has certain limitations in the following areas: (1) Due to the limited compound emotion categories in the DH-FaceEmoVid-150 dataset, some compound emotions are not well represented. For example, the "sadly disgusted" emotion does not successfully blend the two emotions but instead emphasizes the "disgusted" emotion. (2) In terms of fine-grained expression control, our model is trained solely on a combination of action units extracted from real talking videos. This dependency between action units may limit its ability to precisely control a disentangled single action unit. Our future efforts will aim to address these issues.
\section{Conclusion and Future Work}

In this paper, we introduce MoEE, a novel multimodal-guided framework for coarse-grained emotion control and fine-grained expression control in avatar generation, using end-to-end diffusion models, which significantly enhancing controllability and vividness compared to previous models. We develop a mixture of emotion expert module and a emotional talking head dataset, to achieve accurate and vivid controls. Additionally, our Emotion-to-Latents module enables multi-modal emotion control and improves emotion control based solely on audio signals. Experimental results demonstrate MoEE's exceptional lip-sync quality, fine-grained emotion controllability, and the naturalness of the generated outputs. We hope our work will inspire further research into emotional talking heads and encourage more studies in this area.
\vspace{2mm}
 \small \bibliographystyle{ieeenat_fullname} \bibliography{main}

\newpage

\setcounter{section}{0}
\renewcommand\thesection{\Alph{section}}

In this Appendix, we first detail the Mixture of Emotion Experts module and the Emotion-to-Latents module (Appendix \ref{app:network}). Next, we introduce the training and testing details (Appendix \ref{app:training}). Finally, we provide more visualization results (Appendix \ref{app:Visualization}) and further visualizations of the dataset (Appendix \ref{app:Dataset}).

\section{The Networks Details}\label{app:network}
\noindent\textbf{Mixture of Emotion Experts}
In our model, we employ the Mixture of Emotion Experts (MoEE) framework based on cross-attention. Different emotion experts are guided by emotional signals extracted from audio and text to generate single basic emotions. The emotion latent obtained from the Emotion-to-Latents module serves as the key and value in the cross-attention, with a dimension of $(bs, t_{emotion}, c_{emotion})$. Here, bs represents the batch size, $t_{emotion}$ denotes the number of tokens (set to 8), and $c_{emotion}$ refers to the channel dimension of the emotion latent (set to 512). The hidden state is used as the query, which is normalized using LayerNorm before entering the cross-attention module. Finally, a skip connection is applied to prevent issues such as gradient vanishing and gradient explosion, thereby accelerating the convergence of the model.
\noindent\textbf{Emotion-to-Latents}
To achieve control signals across multiple modalities, we introduce the Emotion-to-Latents module. Specifically, we first encode text, label, and audio inputs using pretrained encoders. To map signals from different modalities into a unified emotion latent space, we incorporate a cross-attention module. Four separate fully connected networks are trained to project the channel dimensions of different embeddings into 512 dimensions, which are used as queries. We also define learnable embeddings that serve as keys and values for attention computation. These embeddings, with a channel dimension of 768, are used to compute a new feature representation based on the learnable embeddings. The resulting emotion latents obtained from the cross-attention module are then fed into the UNet for further processing.

\section{Training and Testing Details}\label{app:training}
Experiments for both training and inference were conducted using a platform with 8 NVIDIA A800 GPUs. Each of the two training stages consisted of 30,000 steps, utilizing a batch size of 4 and video dimensions of 512 × 512 pixels. The AdamW \cite{loshchilov2017decoupled} optimizer is employed with a learning rate of 1e-5, and the motion module was initialized using pretrained weights from Animatediff. Each training instance in the second stage produced 14 video frames, with the motion module’s latents concatenated with the first 2 ground truth frames for video continuity. For the diffusion model, a quadratic $\beta$ schedule is set with $\beta_{min}$ = 0.05 and $\beta_{max}$ = 20.

In inference, the model follows the DDIM \cite{song2020denoising} approach and samples 150 steps. Continuity across sequences are ensured by concatenating noisy latents with feature maps of the last 2 motion frames from the previous step within the motion module.

\section{More Visualization Results}\label{app:Visualization}
\begin{figure}[t]  
\centering
\includegraphics[width=0.9\linewidth]{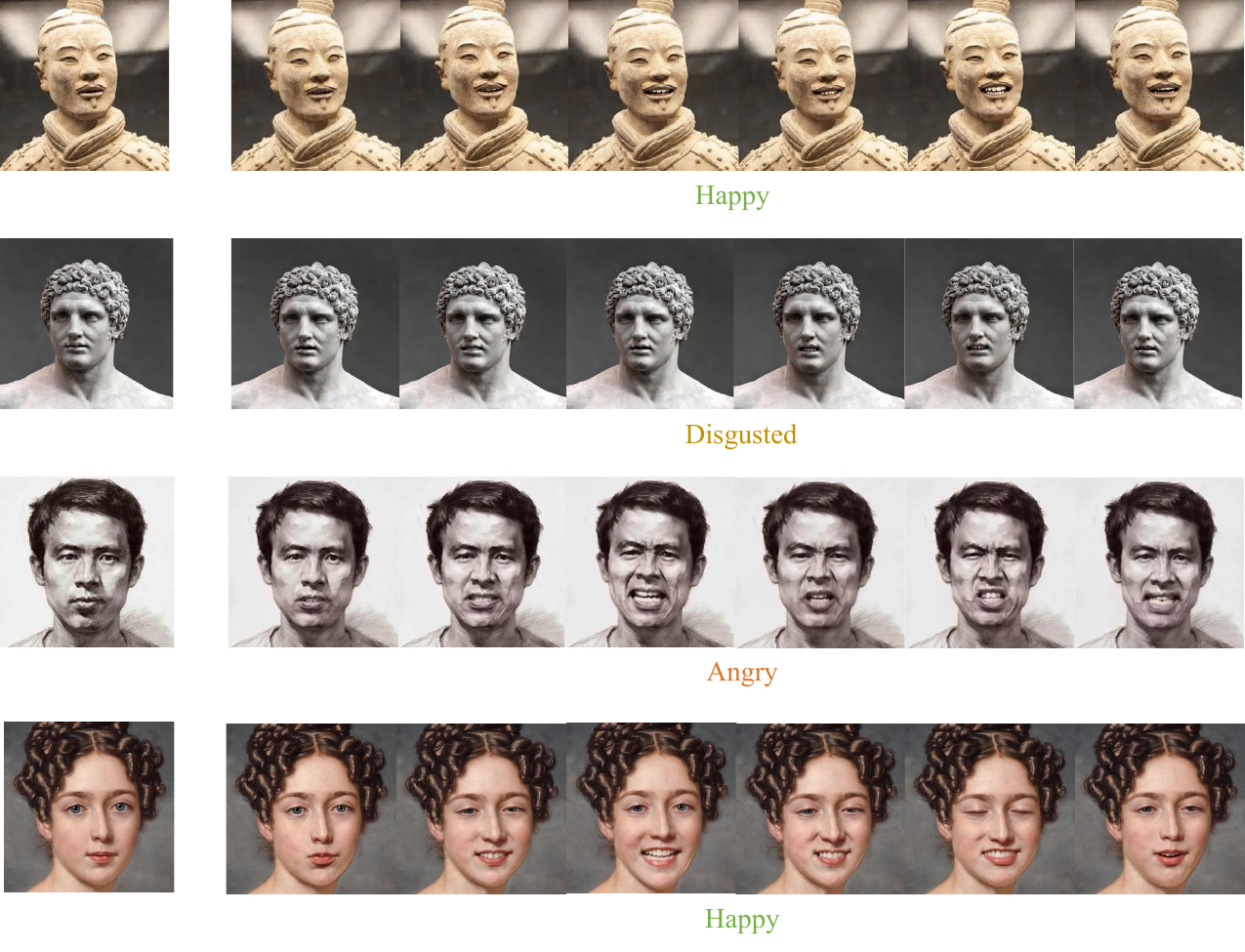}  
\caption{Talking head videos with different portrait styles under various emotions.}
    \label{fig:style_showcase}
    \vspace{-2mm}
\end{figure}

\begin{figure*}[t]  
\centering
\includegraphics[width=0.9\linewidth]{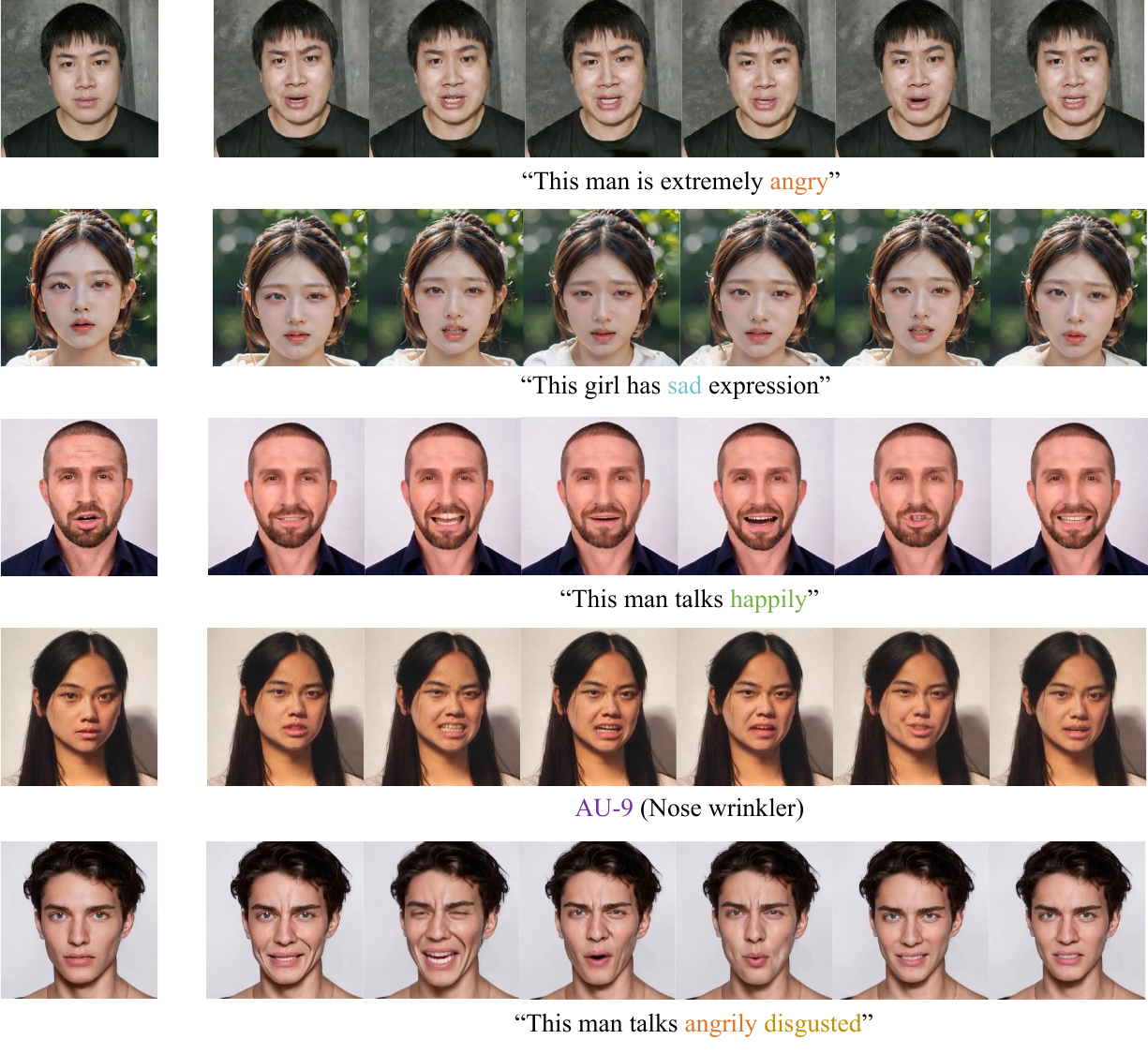}  
\caption{Portrait image animation results on Emotion Control and AU Control.}
    \label{fig:showcase}
    \vspace{-2mm}
\end{figure*}

To further support the conclusions drawn in the main paper, we provide additional results in this section. Figure~\ref{fig:style_showcase} shows more video generation results of the proposed approach with different portrait styles. Figure~\ref{fig:showcase} presents additional examples of the single emotion, compound emotion, and AU controls.

\section{More Visualization of the Dataset}\label{app:Dataset}
\begin{figure*}[t]  
\centering
\includegraphics[width=0.9\linewidth]{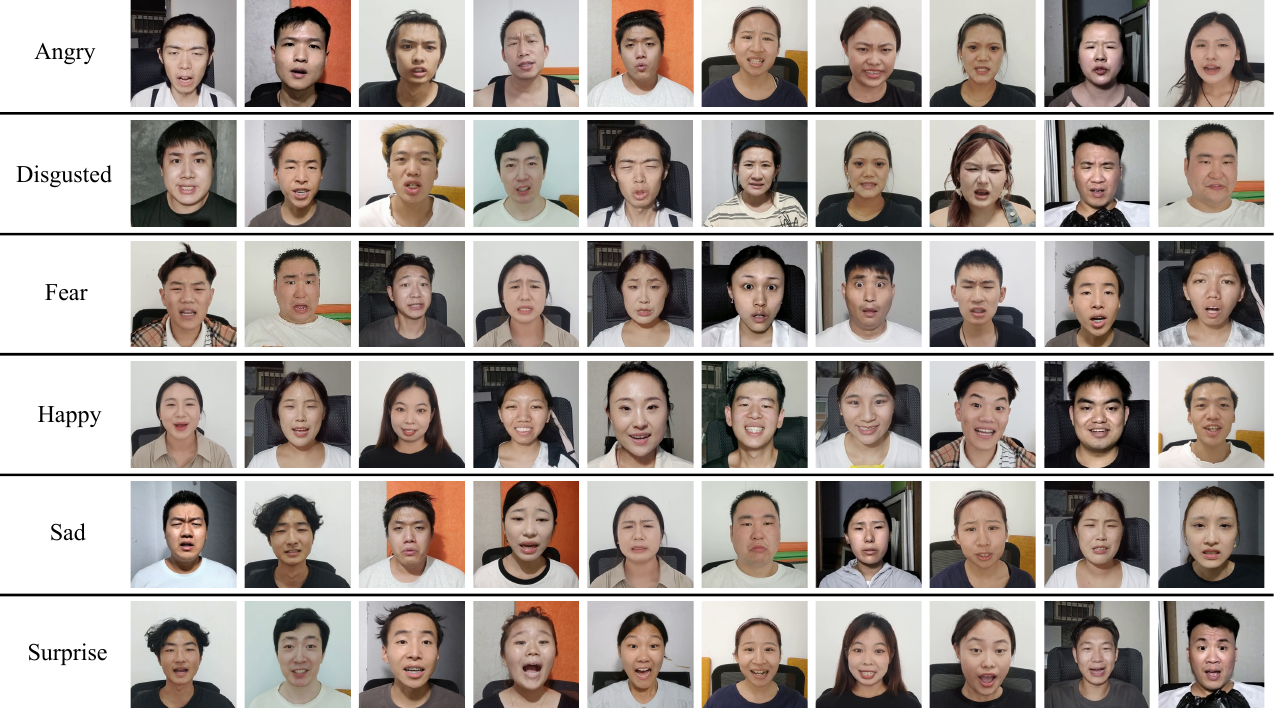}  
\caption{Visualization of basic emotion examples from the DH-FaceEmoVid-150 dataset.}
    \label{fig:dataset_faceemovid_single}
    \vspace{-2mm}
\end{figure*}

\begin{figure*}[t]  
\centering
\includegraphics[width=0.9\linewidth]{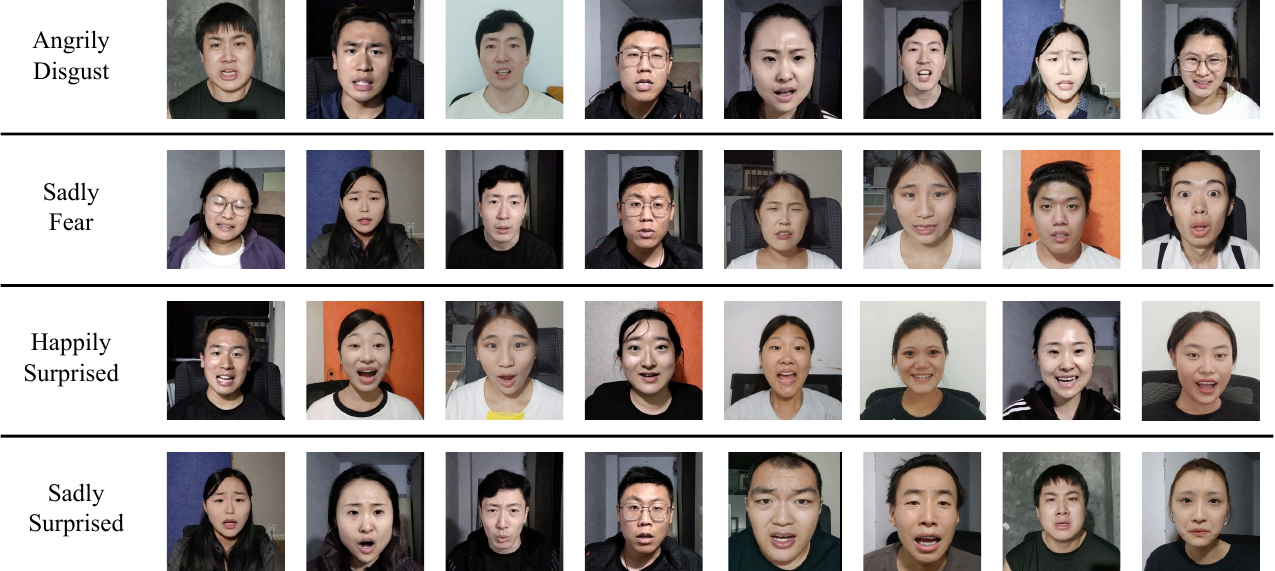}  
\caption{Visualization of compound emotion examples from the DH-FaceEmoVid-150 dataset.}
    \label{fig:dataset_faceemovid_compound}
    \vspace{-2mm}
\end{figure*}

We present additional samples from the DH-FaceEmoVid-150 dataset, including six basic emotions in Figure~\ref{fig:dataset_faceemovid_single} and four compound emotions in Figure~\ref{fig:dataset_faceemovid_compound}. The FPS of the videos in the dataset has been standardized to 30, with each video clip having a duration of 30 seconds.

\end{document}